%% file: paper.tex
\documentclass[runningheads]{llncs}
\usepackage{amsmath,bm}
\usepackage{graphicx}
\usepackage{color}
\usepackage{multirow}
\usepackage{subfigure}

\begin{document}

\title{A Layer-Wise Information Reinforcement Approach to Improve Learning in Deep Belief Networks}
%
\titlerunning{A Layer Wise Information Reinforcement in Deep Belief Networks}
\author{Mateus Roder\inst{1,2}\orcidID{0000-0002-3112-5290} \and
Leandro A. Passos\inst{1,2}\orcidID{0000-0003-3529-3109} \and
Luiz Carlos Felix Ribeiro\inst{1,2}\orcidID{0000-0003-1265-0273} \and
Clayton Pereira\inst{1,2}\orcidID{0000-0002-0427-4880} \and
Jo\~ao Paulo Papa\inst{1,2}\orcidID{0000-0002-6494-7514}\\}
\institute{S\~ao Paulo State University - UNESP, Bauru, Brasil
	\email{\{mateus.roder, clayton.pereira, leandro.passos, luiz.felix, joao.papa\}@unesp.br}\\
	\url{https://www.fc.unesp.br/} \and
	Recogna Laboratory - www.recogna.tech}

\authorrunning{M. Roder et al.}

\maketitle              
\begin{abstract}
With the advent of deep learning, the number of works proposing new methods or improving existent ones has grown exponentially in the last years. In this scenario, ``very deep'' models were emerging, once they were expected to extract more intrinsic and abstract features while supporting a better performance. However, such models suffer from the gradient vanishing problem, i.e., backpropagation values become too close to zero in their shallower layers, ultimately causing learning to stagnate. Such an issue was overcome in the context of convolution neural networks by creating ``shortcut connections'' between layers, in a so-called deep residual learning framework. Nonetheless, a very popular deep learning technique called Deep Belief Network still suffers from gradient vanishing when dealing with discriminative tasks. Therefore, this paper proposes the Residual Deep Belief Network, which considers the information reinforcement layer-by-layer to improve the feature extraction and knowledge retaining, that support better discriminative performance. Experiments conducted over three public datasets demonstrate its robustness concerning the task of binary image classification.
\end{abstract}

\input{sections/introduction.tex}
\input{sections/background.tex}

\input{sections/resdbn.tex}
\input{sections/methodology.tex}
\input{sections/experiments.tex}
\input{sections/conclusions.tex}

\section*{Acknowledgments} The authors are grateful to FAPESP grants \#2013/07375-0, \#2014/12236-1, \#2017/25908-6, \#2019/07825-1, and \#2019/07665-4, as well as CNPq grants \#307066/2017-7, and \#427968/2018-6.

\bibliographystyle{splncs04}
\bibliography{references.bib}

\end{document}

%% file: sections/introduction.tex
\section{Introduction}
\label{s.introduction}

Machine learning-based approaches have been massively studied and applied to daily tasks in the last decades, mostly due to the remarkable accomplishments achieved by deep learning models. Despite the success attained by these techniques, they still suffer from a well-known drawback regarding the backpropagation-based learning procedure: the vanishing gradient. This kind of problem becomes more prominent on deeper models since the gradient vanishes and is not propagated adequately to former layers, thus, preventing a proper parameter update.

To tackle such an issue, He et al.~\cite{he2016deep} proposed the ResNet, a framework where the layers learn residual functions concerning the layer inputs, instead of learning unreferenced functions. In short, the idea is mapping a set of stacked layers to a residual map, which comprises a combination of the set input and output and then mapping it back to the desired underlying mapping.

The model achieved fast popularity, being applied in a wide range of applications, such as traffic surveillance~\cite{jung:17}, medicine~\cite{passosSACI:2018a,khojastehCBM:2019}, and action recognition~\cite{feichtenhofer:16}, to cite a few. Moreover, many works proposed different approaches using the idea of residual functions. Lin et al.~\cite{lin:17}, for instance, proposed the RetinaNet, a pyramidal-shaped network that employs residual stages to deal with one-shot small object detection over unbalanced datasets. Meanwhile, Szegedy et al.~\cite{szegedy:17} proposed the Inception-ResNet for object recognition. Later, Santos et al.~\cite{santos:19video} proposed the Cascade Residual Convolutional Neural Network for video segmentation.

In the context of deep neural networks, there exist another class of methods that are composed of Restricted Boltzmann Machines (RBMs)~\cite{Hinton:02}, a stochastic approach represented by a bipartite graph whose training is given by the minimization of the energy between a visible and a latent layer. Among these methods, Deep Belief Networks (DBNs)~\cite{Hinton:06} and Deep Boltzmann Machines~\cite{salakhutdinov2009deep,passosASOC:19} achieved a considerable popularity in the last years due the satisfactory results over a wide variety of applications~\cite{pereiraCAIP:2017,hassan2019human,wang2019deep}. 

However, as far as we are concerned, no work addressed the concept of reinforcing the feature extraction over those models in a layer-by-layer fashion. Therefore, the main contributions of this paper are twofold: (i) to propose the Residual Deep Belief Network (Res-DBN), a novel approach that combines each layer input and output to reinforce the information conveyed through it, and (ii) to support the literature concerning both DBNs and residual-based models.

The remainder of this paper is presented as follows: Section~\ref{s.background} introduces the main concepts regarding RBMs and DBNs, while Section~\ref{s.resdbn} proposes the Residual Deep Belief Network. Further, Section~\ref{s.methodology} describes the methodology and datasets employed in this work. Finally, Sections~\ref{s.experiments} and~\ref{s.conclusions} provide the experimental results and conclusions, respectively.

%% file: sections/background.tex
\section{Theoretical Background}
\label{s.background}

This section introduces a brief theoretical background regarding Restricted Boltzmann Machines and Deep Belief Networks.

\input{sections/rbm}
\input{sections/dbn}

%% file: sections/rbm.tex
\subsection{Restricted Boltzmann Machines}
\label{ss.rbm}

Restricted Boltzmann Machine stands for a stochastic physics-inspired computational model capable of learning data distribution intrinsic patterns. The process is represented as a bipartite graph where the data composes a visible input-like layer $\bm{v}$, and a latent $n$-dimensional vector $\bm{h}$, composed of a set of hidden neurons whose the model tries to map such inputs onto. The model's training procedure dwells on the minimization of the system's energy, given as follows:

\begin{equation}
 \label{e.energy}
 E(\bm{v},\bm{h}) = -\sum_{i=1}^mb_iv_i-\sum_{j=1}^nc_jh_j-\sum_{i=1}^m\sum_{j=1}^nw_{ij}v_ih_j,
\end{equation}
where $m$ and $n$ stand for the dimensions of the visible and hidden layers, respectively, while $b$ and $c$ denote their respective bias vectors, further, $\textbf{W}$ corresponds to the weight matrix connecting both layers, in which $w_{ij}$ stands for the connection between visible unit $i$ and the $j$ hidden one. Notice the model is restricted, thus implying no connection is allowed among the same layer neurons. 

Ideally, the model was supposed to be solved by computing the joint probability of the visible and hidden neurons in an analytic fashion. However, such an approach is intractable since it requires the partition function calculation, i.e., computing every possible configuration of the system. Therefore, Hinton proposed the Contrastive Divergence (CD)~\cite{Hinton:02}, an alternative method to estimate the conditional probabilities of the visible and hidden neurons using Gibbs sampling over a Monte Carlo Markov Chain (MCMC). Hence, the probabilities of both input and hidden units are computed as follows:

\begin{equation}
\label{e.ph}
p(h_j=1|\bm{v}) = \sigma\left(c_j + \sum_{i=1}^mw_{ij}v_i\right),
\end{equation}
and
\begin{equation}
\label{e.pv}
p(v_i=1|\bm{h}) = \sigma\left(b_i + \sum_{j=1}^nw_{ij}h_j\right),
\end{equation}
where $\sigma$ stands for the logistic-sigmoid function.

%% file: sections/dbn.tex
\subsection{Deep Belief Networks}
\label{ss.dbn}

\begin{figure}[!ht]
 	\centering
  	\caption{DBN architecture with two hidden layers for classification purposes.} 
 	\includegraphics[width=0.36\linewidth]{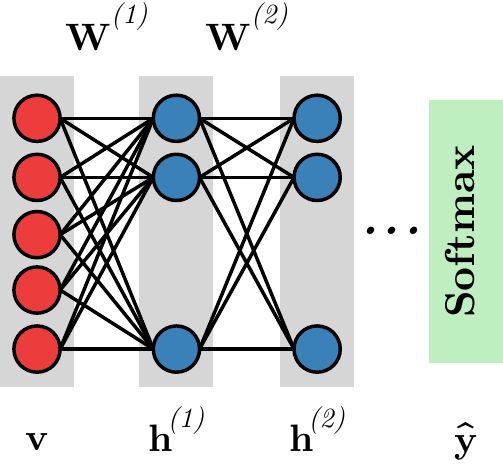}
 	\label{f.dbn}
 \end{figure}

Conceptually, Deep Belief Networks are graph-based generative models composed of a visible and a set of hidden layers connected by weight matrices, with no connection between neurons in the same layer. In practice, the model comprises a set of stacked RBMs whose hidden layers greedily feeds the subsequent RBM visible layer. Finally, a softmax layer is attached at the top of the model, and the weights are fine-tuned using backpropagation for classification purposes. Figure~\ref{f.dbn} depicts the model. Notice that $\textbf{W}^{(l)}$, $l \in [1,L]$, stands for the weight matrix at layer $l$, where $L$ denotes the number of hidden layers. Moreover, $\bm{v}$ stands for the visible layer, as well as $\bm{h}^{(l)}$ represents the $l^{th}$ hidden layer.

%% file: sections/resdbn.tex
\section{Information Reinforcement in DBNs}
\label{s.resdbn}
In this section, we present the proposed approach concerning the residual reinforcement layer-by-layer in Deep Belief Networks, from now on called Res-DBN. Since such a network is a hybrid model between sigmoid belief networks and binary RBMs~\cite{Hinton:06}, it is important to highlight some ``tricks" to make use of the information provided layer-by-layer.

As aforementioned, DBNs can be viewed as hybrid networks that model the data's prior distribution in a layer-by-layer fashion to improve the lower bound from model distribution. Such a fact motivated us to make use of the information learned in each stack of RBM for reinforcement since the greedy-layer pre-training uses the activation of latent binary variables as the input of the next visible layer. Generally speaking, such activation is defined by Eq.~\ref{e.ph}, and its pre-activation vector, $\bm{a}^{(l)}$, as follows:
\begin{equation}
a^{(l)}_j = c^{(l)}_{j} + \sum_{i=1}^mw^{(l)}_{ij}x^{(l-1)}_{i},
\label{e.act}
\end{equation}
where, $c^{(l)}_{j}$ stands for the bias from hidden layer $l$, $m$ is the number of units present on the previous layer, $w^{(l)}_{ij}$ represents the weight matrix for layer $l$, and $x^{(l-1)}_{i}$ stands for the input data from layer $l-1$, where $x^{0}_{i} = v_{i}$. 

Therefore, it is possible to use the ``reinforcement pre-activation'' vector, denoted as $\hat{\bm{a}}^{(l)}$, from layer $l$, $\bm{\forall}$ $l > 1$. Since the standard RBM output of post-activation (provided by Eq.~\ref{e.ph}) is in $[0,1]$ interval, it is necessary to limit the reinforcement term of the proposed approach as follows:
\begin{equation}
\hat{\bm{a}}^{(l)} = \dfrac{\delta(\bm{a}^{(l-1)})} {max\{\delta(a^{(l-1)}_{j})\}},
\label{e.res1}
\end{equation}
where, $\delta$ stands for the Rectifier\footnote{$\delta(z) = max(0, z)$.} function, while $max$ returns the maximum value from the $\delta$ output vector for normalization purposes. Then, the new input data and the information aggregation for layer $l$ is defined by adding the values obtained from Eq.~\ref{e.res1} to the post-activation, i.e., applying $\sigma(\bm{a}^{(l-1)})$, as follows:

\begin{equation}
	x^{(l-1)}_i = \sigma(a^{(l-1)}_j) + \hat{a}^{(l)}_j,
	\label{e.res2}
\end{equation}
where $x^{(l-1)}_i$ stands for the new input data to layer $l$, $\bm{\forall}$ $l > 1$, and its normalized and vectorized form can be obtained as follows:

\begin{equation}
\bm{x}^{(l-1)} = \dfrac{\bm{x}^{(l-1)}}{max\{x^{(l-1)}_i\}} .
\label{e.res3}
\end{equation}

It is important to highlight that, in Eq.~\ref{e.res1}, we only use the positive pre-activations to retrieve and propagate the signal that is meaningful for neurons excitation, i.e., values greater than 0, which generates a probability of more than $50\%$ after applying sigmoid activation.

\begin{figure}[!ht]
	\centering
	\caption{Res-DBN architecture with 3 hidden layers.} 
	\includegraphics[width=0.7\textwidth]{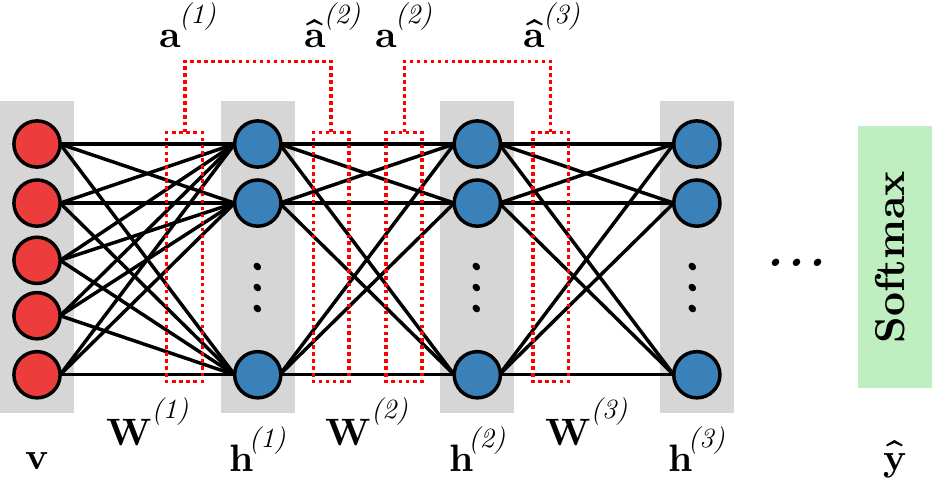}
	\label{f.res_dbn}
\end{figure}

The Figure~\ref{f.res_dbn} depicts the Res-DBN architecture, with hidden layers connected by the weights $\textbf{W}^{(l)}$. The dashed connections stand for the reinforcement approach, with the information aggregation occuring as covered by the Eqs.~\ref{e.act} to \ref{e.res3}, from a generic hidden layer to the next one ($\bm{h}^{(1)} \rightarrow \bm{h}^{(2)}$, for instance).
 

%% file: sections/methodology.tex
\section{Methodology}
\label{s.methodology}

In this section, we present details regarding the datasets employed in our experiments, as well as the experimental setup applied for this paper. 
\subsection{Datasets}
\label{ss.datasets}

Three well-known image datasets were employed throughout the experiments:

\begin{itemize}
\item[$\bullet$]MNIST\footnote{http://yann.lecun.com/exdb/mnist}~\cite{Lecun:98}: set of $28 \times 28$ binary images of handwritten digits (0-9), i.e., 10 classes. The original version contains a training set with $60,000$ images from digits `0'-`9', as well as a test set with $10,000$ images.
	
\item[$\bullet$]Fashion-MNIST\footnote{https://github.com/zalandoresearch/fashion-mnist}~\cite{Xiao:17}: set of $28 \times 28$ binary images of clothing objects. The original version contains a training set with $60,000$ images from $10$ distinct objects (t-shirt, trouser, pullover, dress, coat, sandal, shirt, sneaker, bag, and ankle boot), and a test set with $10,000$ images.

\item[$\bullet$]Kuzushiji-MNIST\footnote{https://github.com/rois-codh/kmnist}~\cite{Clanuwat:18}: set of $28 \times 28$ binary images of hiragana characters. The original version contains a training set with $60,000$ images from $10$ previously selected hiragana characters, and a test set with $10,000$ images.
\end{itemize}

\subsection{Experimental Setup}
\label{ss.setup}

Concerning the experiments, we employed the concepts mentioned in Section~\ref{s.resdbn}, considering two main phases: (i) the DBN pre-training and (ii) the discriminative fine-tuning. Regarding the former, it is important to highlight that the information reinforcement is performed during the greedy layer-wise process, in which the hidden layers ($l = {1, 2, \dots, L}$) receive the positive ``residual'' information. Such a process takes into account a mini-batch of size $128$, a learning rate of $0.1$, $50$ epochs for the bottommost RBM convergence, and $25$ epochs for the intermediate and top layers convergence\footnote{Such a value is half of the initial one to evaluate Res-DBN earlier convergence.}.

Moreover, regarding the classification phase, a softmax layer was attached at the top of the model after the DBN pre-training, performing the fine-tuning process for $20$ epochs through backpropagation using the well-known ADAM~\cite{kingma2014adam} optimizer. The process employed a learning rate of $10^{-3}$ for all layers. Furthermore, it was performed $15$ independent executions for each model to provide statistical analysis. To assess the robustness of the proposed approach, we employed seven different DBN architectures changing the number of hidden neurons and layers, as denoted in Table~\ref{t.models}.


\begin{table}[]
	\renewcommand{\arraystretch}{1}
	\centering
		\begin{tabular}{llcl}
			\hline
				\textbf{Model} &
				\multicolumn{1}{c}{\textbf{Res-DBN}} &
				&
				\multicolumn{1}{c}{\textbf{DBN}} \\
			\hline
			(a) & $i$:500:500:10 &~& $i$:500:500:10\\
			(b) & $i$:500:500:500:10 &~& $i$:500:500:500:10\\
			(c) & $i$:500:500:500:500:10 &~& $i$:500:500:500:500:10\\
			(d) & $i$:1000:1000:10 &~& $i$:1000:1000:10\\
			(e) & $i$:1000:1000:1000:10 &~& $i$:1000:1000:1000:10\\
			(f) & $i$:1000:1000:1000:1000:10 &~& $i$:1000:1000:1000:1000:10\\
			(g) & $i$:2000:2000:2000:2000:10 &~& $i$:2000:2000:2000:2000:10\\
			\hline
		\end{tabular}
	\\~\\
	\caption{Different setups, where $i$ stands for the number of neurons on the input layer.}
	\label{t.models}
\end{table}

%% file: sections/experiments.tex
\section{Experiments}
\label{s.experiments}

In this Section, we present the experimental results concerning seven distinct DBN architectures, i.e., (a), (b), (c), (d), (e), (f) and (g), over the aforementioned datasets. Table~\ref{t.results} provides the average accuracies and standard deviations for each configuration on $15$ trials, where the proposed approach is compared against the standard DBN formulation in each dataset for each configuration. Further, results in bold represent the best values according to the statistical Wilcoxon signed-rank test~\cite{Wilcoxon:45} with significance $p\leq 0.05$ concerning each model configuration. On the other hand, underlined values represent the best results overall models regarding each dataset, without a statistical difference, i.e., results similar to the best one achieved.

\begin{table}[]
	\renewcommand{\arraystretch}{1.4}
	\centering
	\resizebox{\textwidth}{!}{
	\begin{tabular}{lccccccccccccc}
		\hline
			\multirow{2}{*}{\textbf{Experiment}} &
			\multicolumn{3}{c}{\textbf{MNIST}}  & & &
			\multicolumn{3}{c}{\textbf{Fashion MNIST}} & & &
			\multicolumn{3}{c}{\textbf{Kuzushiji MNIST}}
		\\
			& Res-DBN
			& & DBN
			& & & Res-DBN
			& & DBN
			& & & Res-DBN
			& & DBN
		\\
		\hline
		(a)
			& $\bm{97.39 \pm 0.08}$ &~
			& $97.23 \pm 0.09$ &~&~

			& $81.13 \pm 0.33$ &~
			& $\bm{81.52 \pm 0.27}$ &~&~

			& $\bm{86.49 \pm 0.18}$ &~ 
			& $84.78 \pm 0.29$ \\
		(b)
			& $\bm{97.61 \pm 0.07}$ &~
			& $97.44 \pm 0.11$ &~&~

			& $81.49 \pm 0.50$  &~
			& $81.41 \pm 0.57$ &~&~

			& $\bm{87.75 \pm 0.20}$ &~ 
			& $85.81 \pm 0.18$ \\
		(c)
			& $97.59 \pm 0.10$ &~
			& $97.57 \pm 0.09$ &~&~

			& $81.66 \pm 0.33$ &~
			& $81.51 \pm 0.60$ &~&~

			& $\bm{88.21 \pm 0.18}$ &~ 
			& $86.97 \pm 0.30$ \\
		(d) 
			& $\bm{97.66 \pm 0.10}$ &~
			& $97.40 \pm 0.10$ &~&~

			& $81.55 \pm 0.35$ &~
			& $81.15 \pm 0.64$ &~&~

			& $\bm{87.67 \pm 0.19}$ &~ 
			& $86.24 \pm 0.21$ \\
		(e) 
			& $\underline{\bm{97.85 \pm 0.06}}$ &~
			& $97.48 \pm 0.12$ &~&~

			& $\bm{82.05 \pm 0.48}$ &~
			& $81.59 \pm 0.51$ &~&~

			& $\bm{88.95 \pm 0.16}$ &~ 
			& $87.57 \pm 0.20$ \\
		(f) 
			& $\underline{97.80 \pm 0.37}$ &~
			& $97.68 \pm 0.29$ &~&~

			& $82.16 \pm 0.50$  &~
			& $82.19 \pm 0.46$ &~&~

			& $\underline{\bm{89.63 \pm 0.23}}$ &~ 
			& $88.81 \pm 0.40$ \\
		(g) 
			& $\underline{\bm{97.88 \pm 0.19}}$ &~
			& $97.51 \pm 0.30$ &~&~

			& $\underline{82.73 \pm 0.53}$ &~
			& $\underline{82.63 \pm 0.36}$ &~&~
			
			& $\underline{\bm{89.45 \pm 0.78}}$ &~ 
			& $88.70 \pm 0.60$ \\
		\hline
	\end{tabular}}
	\\~\\
	\caption{Experimental results on different datasets.}
	\label{t.results}
\end{table}

Regarding the original MNIST dataset, the preeminence of the proposed model over the standard version of the RBM is evident, since the best results were obtained exclusively by Res-DBN and, from these, five out of seven scenarios presented statistical significance. Such a behavior is stressed in the Kuzushiji MNIST dataset, where the best results were obtained solely by the Res-DBN over every possible configuration. The results' similarity between these datasets is somehow expected since both are composed of handwritten digits or letters.



The Fashion MNIST dataset presents the single experimental scenario, i.e., model (a), where the proposed model was outperformed by the traditional DBN, although by a small margin. In all other cases Res-DBN presented results superior or equal to the traditional formulation, which favors the Res-DBN use over the DBNs.

Finally, one can observe the best results overall were obtained using a more complex model, i.e., with a higher number of layers and neurons, as denoted by the underlined values. Additionally, the proposed model outperformed or at least is equivalent, to the standard DBN in virtually all scenarios, except one concerning the Fashion-MNIST dataset.

\subsection{Training Evaluation}
\label{ss.trainingEvaluation}

\begin{figure}
\centering
\caption{Accuracy on MNIST test set.}
\includegraphics[width=0.9\textwidth]{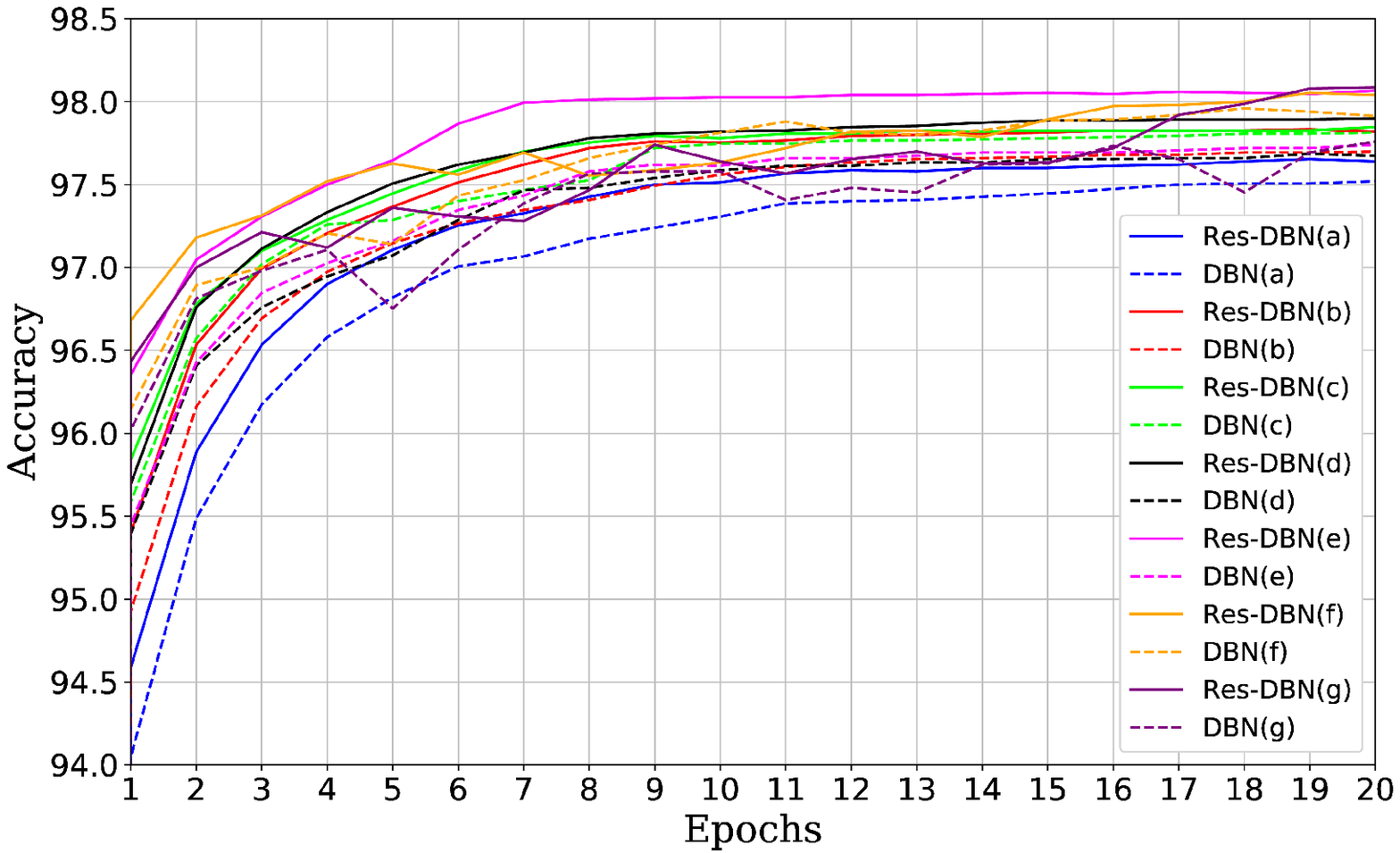}
\label{f.mnist}
\end{figure}

\begin{figure}
\centering
\caption{Accuracy on Fashion MNIST test set.}
\includegraphics[width=0.9\textwidth]{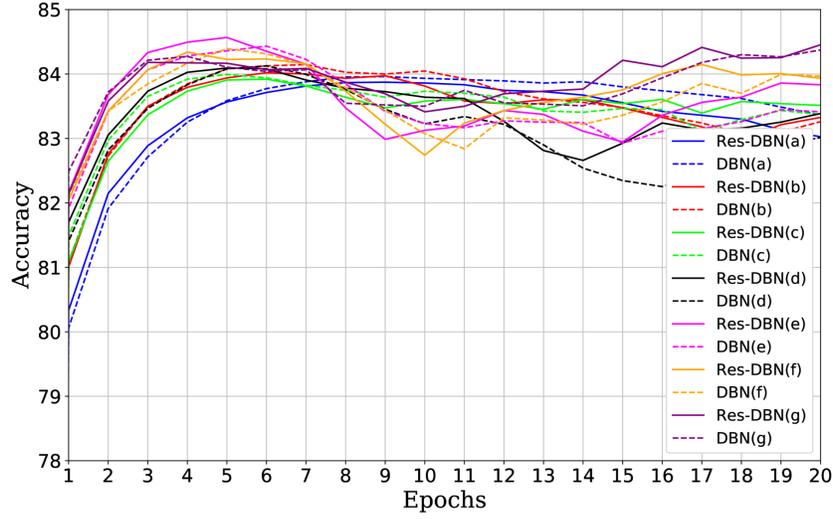}
\label{f.fash}
\end{figure}

\begin{figure}
\centering
\caption{Accuracy on Kuzushiji MNIST test set.}
\includegraphics[width=0.9\textwidth]{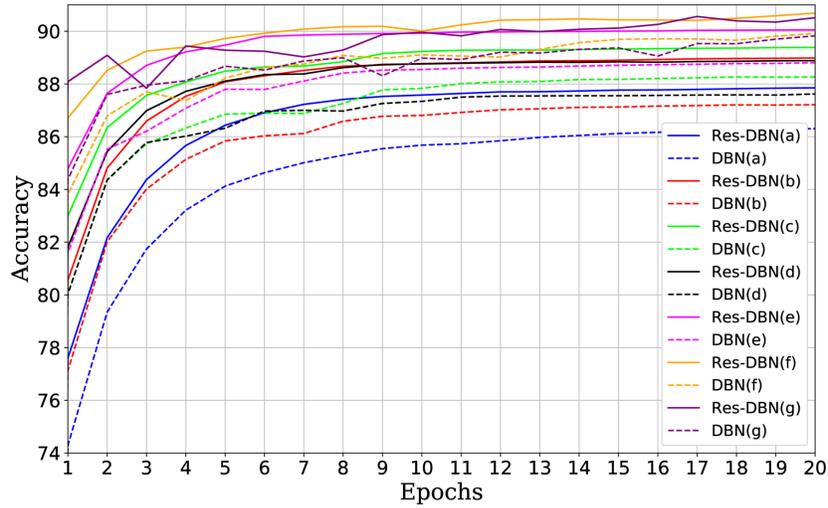}
\label{f.kmnist}
\end{figure}

Figures~\ref{f.mnist},~\ref{f.fash}, and~\ref{f.kmnist} depict the models' learning curves over the test sets regarding MNIST, Fashion MNIST, and Kuzushiji MNIST, respectively. In Figure~\ref{f.mnist}, one can observe that Res-DBN(e) converged faster than the remaining approaches, obtained reasonably good results after seven iterations. At the end of the process, Res-DBN(f) and (g) boosted and outperformed Res-DBN(e), as well as any of standard DBN approaches, depicted as dashed lines.

Regarding Fashion MNIST, it can be observed in Figure~\ref{f.fash} that Res-DBN(e) was once again the fastest technique to converge, obtaining acceptable results after five iterations. However, after iteration number five, all models seem to overfit, explaining the performance decrease observed over the testing samples. Finally, after $14$ iterations, the results start increasing once again, being Res-DBN(g) the most accurate technique after $20$ iterations.

Finally, the Kuzushiji learning curve, depicted in Figure~\ref{f.kmnist}, displays a behavior silimiar to the MNIST dataset. Moreover, it shows that Res-DBN provided better results than its traditional variant in all cases right from the beginning of the training. In some cases with a margin greater than $2\%$, showing a promissing improvement.

%% file: sections/conclusions.tex
\section{Conclusions}
\label{s.conclusions}

In this paper, we proposed a novel approach based on reinforcing DBN's layer-by-layer feature extraction in a residual fashion, the so-called Residual Deep Belief Network. Experiments conducted over three public datasets confirm the sturdiness of the model. Moreover, it is important to highlight faster convergence achieved by Res-DBN in front of DBN, once half of the epochs were employed for pre-training hidden layers, and the results outperformed the latter model.

Regarding future work, we intend to investigate the model in the video domain, applying it to classification and recognition tasks, as well as to propose a similar approach regarding Deep Boltzmann Machines.